\documentclass[11pt,a4paper]{article}
\usepackage[hyperref]{emnlp-ijcnlp-2019}
\usepackage{times}
\usepackage{latexsym}
\usepackage{url}
\usepackage{enumitem}
\usepackage{stfloats}
\usepackage{float}
\usepackage{nccmath}
\usepackage{mathtools}
\usepackage{epstopdf}
\usepackage{amsfonts}
\usepackage{amssymb}
\usepackage{CJKutf8}
\usepackage{CJK}
\usepackage{amsmath}
\usepackage{verbatim}
\usepackage{multirow}
\usepackage{subfigure}
\usepackage{pifont}
\usepackage{array}
\usepackage{fp}
\usepackage{makecell}
\usepackage{flushend}
\usepackage{cases}
\usepackage{makecell}
\usepackage{color, soul}
\usepackage{xcolor}
\usepackage{bm}
\usepackage{booktabs}
\usepackage{tablefootnote}
\usepackage{tcolorbox}
\usepackage{algorithm}
\usepackage{algorithmic}
\usepackage{siunitx}
\usepackage{graphicx}
\aclfinalcopy 

\title{Low-Resource Response Generation with Template Prior}

\author{
  Ze Yang$^\dag$, Wei Wu$^\diamondsuit$, Jian Yang$^{\dag}$, Can Xu$^\diamondsuit$, Zhoujun Li$^\dag$\thanks{~~~Corresponding Author}~~~~~\\
  $^\dag$State Key Lab of Software Development Environment, Beihang University, Beijing, China\\
  $^\diamondsuit$Microsoft Corporation, Beijing, China\\
  \{tobey, jiaya, lizj\}@buaa.edu.cn 
  \{wuwei, caxu\}@microsoft.com
}

\date{}

\begin{document}

\maketitle
\begin{abstract}
We study open domain response generation with limited message-response pairs. The problem exists in real-world applications but is less explored by the existing work. Since the paired data now is no longer enough to train a neural generation model, we consider leveraging the large scale of unpaired data that are much easier to obtain, and propose response generation with both paired and unpaired data. The generation model is defined by an encoder-decoder architecture with templates as prior, where the templates are estimated from the unpaired data as a neural hidden semi-markov model. By this means, response generation learned from the small paired data can be aided by the semantic and syntactic knowledge in the large unpaired data. To balance the effect of the prior and the input message to response generation, we propose learning the whole generation model with an adversarial approach. Empirical studies on question response generation and sentiment response generation indicate that when only a few pairs are available, our model can significantly outperform several state-of-the-art response generation models in terms of both automatic and human evaluation. 
\end{abstract}

\section{Introduction}
Human-machine conversation is a long-standing goal of artificial intelligence. Early dialogue systems are designed for task completion with conversations restricted in specific domains \cite{young2013pomdp}. Recently, thanks to the advances in deep learning techniques \cite{sutskever2014sequence,vaswani2017attention} and the availability of large amounts of human conversation on the internet, building an open domain dialogue system with data-driven approaches has become the new fashion in the research of conversational AI.  Such dialogue systems can generate reasonable responses without any needs on rules, and have powered products in the industry such as Amazon Alexa \cite{ram2018conversational} and Microsoft XiaoIce \cite{shum2018eliza}.

State-of-the-art open domain response generation models are based on the encoder-decoder architecture \cite{vinyals2015neural,shangL2015neural}. On the one hand, with proper extensions to the vanilla structure, existing models now are able to naturally handle conversation contexts \cite{serban2015building,xing2017hierarchical}, and synthesize responses with various styles \cite{wang2017steering}, emotions \cite{zhou2017emotional}, and personas \cite{li2016persona}. On the other hand, all the existing success of open domain response generation builds upon an assumption that the large scale of paired data \cite{shao2016generating} or conversation sessions \cite{sordoni2015neural} are available.  In this work, we challenge this assumption by arguing that one cannot always obtain enough pairs or sessions for training a neural generation model.  For example, although it has been indicated by existing work \cite{li2016learning,wang2018learning} that question asking in conversation can enhance user engagement, we find that in a public dataset\footnote{\url{http://tcci.ccf.org.cn/conference/2018/dldoc/trainingdata05.zip}} with $5$ million conversation sessions crawled from Weibo, only $7.3$\% sessions have a question response and thus can be used to learn a question generator for responding\footnote{Questions are detected with the rules in \cite{wang2018learning}.}. When we attempt to generate responses that express positive sentiment, we only get $360$k pairs ($18$\%) with positive responses from a dataset with $2$ million message-response pairs crawled from Twitter. Indeed, existing big conversation data mix various intentions, styles, emotions, personas, and so on. Thus, we have to face the data sparsity problem, as long as we attempt to create a generation model with constraints on responses.

In this work, we jump out of the paradigm of learning from large scale paired data\footnote{The study in this work starts from response generation for single messages. One can easily extend the proposed approach to handle conversation history.}, and investigate how to build a response generation model with only a few pairs at hand. Aside from the paired data, we assume that there are a large number of unpaired data available.  The assumption is reasonable since it is much easier to get questions or sentences with positive sentiment than to get such responses paired with messages. We formalize the problem as low-resource response generation from paired and unpaired data, which is less explored by existing work. Since the paired data are insufficient for learning the mapping from a message to a response, the challenge of the task lies in how to effectively leverage the unpaired data to enhance the learning on the paired data.  Our solution to the challenge is a two-stage approach where we first distill templates from the unpaired data and then use them to guide response generation. Targeting on an unsupervised approach to template learning, we propose representing the templates as a neural hidden semi-markov model (NHSMM) estimated through maximizing the likelihood of the unpaired data. Such latent templates encode both semantics and syntax of the unpaired data and then are used as prior in an encoder-decoder architecture for modeling the paired data.  With the latent templates, the whole model is end-to-end learnable and can perform response generation in an explainable manner. To ensure the relevance of responses regarding input messages and at the same time make full use of the templates, we propose learning the generation model with an adversarial approach. 

Empirical studies are conducted on two tasks: question response generation and sentiment response generation. For the first task, we exploit the dataset published in \cite{wang2018learning} and augment the data with questions crawled from Zhihu\footnote{\url{https://en.wikipedia.org/wiki/Zhihu}}. For the second task, we build a paired dataset from Twitter by filtering responses with an off-the-shelf sentiment classifier and augment the dataset with tweets in positive sentiment extracted from a large scale tweet dataset published in \cite{cheng2010you}. Evaluation results on both automatic metrics and human judgment indicate that with limited message-response pairs, our model can significantly outperform several state-of-the-art response generation models. The source code is available online. \footnote{\url{https://github.com/TobeyYang/S2S_Temp}}

Our contributions in this work are three-folds: (1) proposal of low-resource response generation with paired and unpaired data for open domain dialogue systems; (2) proposal of encoder-decoder with template prior; and (3) empirical verification of the effectiveness of the model with two large-scale datasets. 
   
\section{Related Work}
Inspired by neural machine translation, early work applies the sequence-to-sequence with attention model \cite{shangL2015neural} to open domain response generation, and gets promising results. Later, the basic architecture is extended to suppress generic responses \cite{li2015diversity,zhao2017learning,xing2017topic}; to model the structure of conversation contexts \cite{serban2015building}; and to incorporate different types of knowledge into generation \cite{li2016persona,zhou2017emotional}.  In addition to model design, how to learn a generation model \cite{li2016deep,li2017adversarial}, and how to evaluate the models \cite{liu2016not,lowe2017towards,tao2018ruber}, are drawing attention in the community of open domain dialogue generation. In this work, we study how to learn a response generation model from limited pairs, which breaks the assumption made by existing work. We propose response generation with paired and unpaired data. As far as we know, this is the first work on low-resource response generation for open domain dialogue systems. 

Traditional template-based text generation \cite{becker2002practical,foster2004techniques,gatt2009simplenlg} relies on handcrafted templates that are expensive to obtain. Recently, some work explores how to automatically mine templates from plain text and how to integrate the templates into neural architectures to enhance interpretability of generation. Along this line, \citet{duan-etal-2017-question} mine patterns from related questions in community QA websites and leverage the patterns with a retrieval-based approach and a generation-based approach for question generation. \citet{wiseman2018learning} exploit a hidden semi-markov model for joint template extraction and text generation. In addition to structured templates, raw text retrieved from indexes is also used as ``soft templates'' in various natural language generation tasks \cite{ guu2018generating,pandey2018exemplar,cao2018retrieve,peng2019text}. In this work, we leverage templates for open domain response generation. Our idea is inspired by \cite{wiseman2018learning}, but latent templates estimated from one source are transferred to another source in order to handle the low-resource problem, and the generation model is learned by an adversarial approach rather than by maximum likelihood estimation. 

Before us, the low-resource problem has been studied in tasks such as machine translation \cite{gu-etal-2018-meta,gu-etal-2018-universal}, pos tagging \cite{kann2018character}, word embedding \cite{jiang2018learning}, automatic speech recognition \cite{tuske2014data}, task-oriented dialogue systems \cite{tran2018dual,mi2019meta}, etc. In this work, we pay attention to low-resource open domain response generation which is untouched by existing work. We propose attacking the problem with unpaired data, which is related to the effort in low-resource machine translation with monolingual data \cite{gulcehre2015using,sennrich2015improving,zhang2016exploiting}.  Our method is unique in that rather than using the unpaired data through multi-task learning \cite{zhang2016exploiting} or back-translation \cite{sennrich2015improving}, we extract linguistic knowledge from the data as latent templates and use the templates as prior in generation. 
\section{Low-Resource Response Generation}
\label{sec:generator}
In this section, we first formalize the setting upon which we study low-resource response generation and then elaborate the model of response generation with paired and unpaired data, including how to learn latent templates from the unpaired data, and how to perform generation with the templates.
\subsection{Problem Formalization}
Suppose that we have a dataset $\mathcal{D}_{P}=\{(X_i,Y_i)\}^n_{i=1}$, where $\forall i$, $(X_i, Y_i)$ is a pair of message-response, and $n$ represents the number of pairs in $\mathcal{D}_P$. Different from existing work, we assume that $n$ is small (e.g., a few hundred thousands) and further assume that there is another set $\mathcal{D}_U=\{T_i\}^N_{i=1}$ with $T_i$ a piece of plain text sharing the same characteristics with $\{Y_i\}_{i=1}^n$ (e.g., both are questions) and $N>n$.  Our goal is to learn a generation probability $P(Y|X)$ with both $\mathcal{D}_{P}$ and $\mathcal{D}_U$. Thus, given a new message $X$, we can generate a response $Y$ for $X$ following  $P(Y|X)$. 

Since the limited resource in $\mathcal{D}_P$ may not support accurately learning of $P(Y|X)$, we try to transfer the linguistic knowledge in $\mathcal{D}_U$ to response generation. The challenges then lie in two aspects: (1) how to represent the linguistic knowledge in $\mathcal{D}_U$; and (2) how to effectively leverage the knowledge extracted from $\mathcal{D}_U$ for response generation, given that $\mathcal{D}_U$ cannot provide any information of correspondence between a message $X$ and a response $Y$. The remaining part of the section will describe our solutions to the two problems. 

\subsection{Learning Templates from $\mathcal{D}_U$}
In the representation of the knowledge in $\mathcal{D}_U$, we hope that both semantic information and syntactic information can be kept. Thus, we consider extracting templates from $\mathcal{D}_U$ as the knowledge. A template segments a piece of text as a structured representation. With the templates, semantically and functionally similar text segments are grouped together. Since the templates encode the structure of language in $\mathcal{D}_U$, they can inform the generation model about how to express a response in a desired way (e.g., as a question or with the specific sentiment).   Here, we prefer an unsupervised and parametric approach to learning templates, since ``unsupervised'' means that the approach is generally applicable to various tasks, and ``parameteric'' allows us to naturally incorporate the templates into the generation model. Then, a natural choice for template learning is the neural hidden semi-markov model (NHSMM) \cite{dai2016recurrent,wiseman2018learning}.

NHSMM is an HSMM parameterized with neural networks. HSMM \cite{murphy2002hidden} extends HMM by allowing a hidden state to emit a sequence of observations and thus can segment a piece of text with the latent variables and group similar segments by the variables. Formally, given an observed sequence $Y=(y_1,\ldots,y_S)$, the joint distribution of $Y$ and its segmentation is 
\begin{align*}
    \prod_{t=1}^{S'} P(z_{t+1}, l_{t+1} | z_t, l_t) \prod_{t=1}^{S'} P(y_{i(t-1)+1: i(t)}|z_t,l_t),
\end{align*}
where $z_t\in \{1,\ldots,K\}$ is the hidden state for step $t$, $l_t \in \{1,\ldots, D\}$ is the duration variable for $z_t$ that represents the number of tokens emitted by $z_t$, $i(t)=\sum_{j=1}^t l_j$ with $i(0)=0$ and $i(S')=S$, and $y_{i(t-1)+1: i(t)}$ is the sequence of $(y_{i(t-1)+1}, \ldots, y_{i(t)})$. $P(z_{t+1}, l_{t+1} | z_t, l_t)$ is factorized as $P(z_{t+1}| z_t) \times P(l_{t+1}|z_{t+1})$ where $P(l_{t+1}|z_{t+1})$ is a uniform distribution on $\{1,\ldots, D\}$ and  $P(z_{t+1}| z_t)$ can be viewed as a transition matrix $[A(i,j)]_{K \times K}$ which is defined by 
\begin{align*}
    A(i,j) &\triangleq P(z_{t+1}=j|z_t=i)\\
    &= \frac{\exp(e_j^\top e_i + b_{i,j})}{\sum^K_{o=1} \exp(e_o^\top e_i + b_{i,o})},
\end{align*}
where $e_i, e_j, e_o \in \mathbb{R}^{d_1}$ are embeddings of state $i, j, o$ respectively, and $b_{i,j}, b_{i,o}$ are scalar bias terms. In practice, we set $b_{i,j}=-\infty \Leftrightarrow i=j$ to disable self-transition, because the adjacent states play different syntactic or semantic roles in a desired template. The emission distribution $P(y_{i(t-1)+1: i(t)}|z_t,l_t)$ is defined by
\begin{align*}
    P(y_{i(t-1)+1: i(t)} |z_t,l_t) = P(y_{i(t-1)+1}|z_t,l_t)&\\
    \times \prod_{j=2}^{l_t} P(y_{i(t-1)+j}|y_{i(t-1)+j-1}, z_t,l_t),&
\end{align*}
and parameterized with a recurrent neural network with gated recurrent unit (GRU) \cite{cho-etal-2014-learning}. The hidden vector for position $j$ is formulated as 
\begin{equation}
    \label{gatedstate}
    \begin{split}
        o_j^t =& \text{GRU}\textsubscript{H}(o_{j-1}^t, [e_{z_t}; e_{y_{i(t-1)+j-1}}])\\
        v_j^t =& g_{z_t}\odot o_j^t,
    \end{split}
\end{equation}
where $o_j^t \in \mathbb{R}^{d_2}$, $e_{y_{i(t-1)+j-1}} \in \mathbb{R}^{d_3}$ is the embedding of word $y_{i(t-1)+j-1}$, $[\cdot;\cdot]$ is a concatenation operator, $\odot$ refers to element-wise multiplication,  and $g_{z_t} \in \mathbb{R}^{d_2}$ is a gate (in total, there are $K$ gate vectors as parameters). $P(y_{i(t-1)+j}|y_{i(t-1)+j-1}, z_t,l_t)$ is then defined by
\begin{align*}
    \text{softmax}(W_1 v^t_j + b_1),
\end{align*}
where $W_1 \in \mathbb{R}^{V \times d_2}$ and $b_1\in \mathbb{R}^{d_2}$ are parameters with $V$ the vacabulary size. 

Following \citet{murphy2002hidden}, the marginal distribution of $Y$ can be obtained by the backward algorithm which is formulated as
\begin{equation}
\label{backward}
    \begin{split}
        \beta_t(i) &\triangleq P(y_{t+1:S}|q_t=i) = \sum^K_{j=1}\beta^{*}_t(j)A(i,j)\\
        \beta^{*}_t(j) &\triangleq P(y_{t+1:S}|q_{t+1}=j)\\
                   &=\sum^D_{d=1} \Big [ \beta_{t+d}(j) P(d|j) P(y_{t+1:t+d}|j,d) \Big]\\
        P(Y) &= \sum^K_{j=1}\beta^{*}_0(j)P(q_1=j).
    \end{split}
\end{equation}
where $q_t$ is the hidden state of the $t$-th word in $Y$, and the base cases $\beta_S(i)=1$, $\forall i \in \{1,\ldots,K\}$. Specifically, to learn more reasonable segmentations, we parsed every sentence by stanford parser \cite{manning-EtAl:2014:P14-5} and forced NHSMM not to break syntactic elements such as VP and NP, etc. The parameters of the NHSMM are estimated by maximizing the log-likelihood of $\mathcal{D}_U$ through backpropagation. 

\begin{figure*}[ht!]
    \centering
    \includegraphics[width=\linewidth]{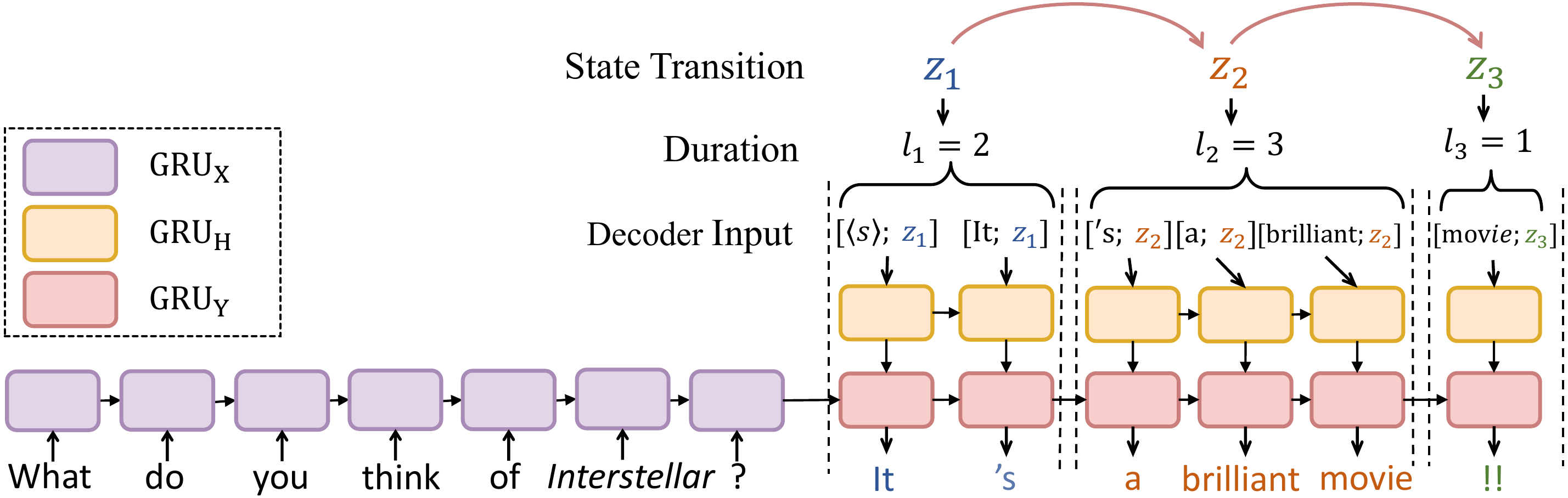}
    \caption{The architecture of the generation model.}
    \label{fig:generator}
\end{figure*}

\subsection{Response Generation with Template Prior} \label{sec: gen_prior}

We propose incorporating the templates parameterized by the NHSMM learned from $\mathcal{D}_U$ into response generation as prior. Figure \ref{fig:generator} illustrates the architecture of the generation model.  In a nutshell, the model first samples a chain of states with duration as a template. The template specifies a segmentation of the response to generate. Then, the hidden representations of the segments 
defined by Equation (\ref{gatedstate}) are fed to an encoder-decoder architecture for response generation, where the hidden states of the decoder are calculated with both attention over the hidden states of the input message given by the encoder and the hidden representations of the segments given by the template prior. The template prior acts as a base and assists the encoder-decoder in response generation regarding to an input message, when paired information is insufficient for learning the correspondence between a message and a response.   Note that similar to the conditional variational autoencoder (CVAE) \cite{zhao2017learning}, our model also exploits hidden variables for response generation. The difference is that the hidden variables in our model are structured and learned from extra resources, and thus encode more semantic and syntactic information.

Specifically, we segment responses in $\mathcal{D}_U$ with Viterbi algorithm \cite{zucchini2016hidden}, collect all chains of states as a pool $\mathcal{P}$ and sample a chain from the pool uniformly. We do not sample states according to the transition matrix $[A(i,j)]_{K \times K}$, since it is difficult to determine the end of a chain. Suppose that the sampled chain is $(z_1,\ldots z_S)$, then $\forall 1\leq t \leq S$, we sample an $l_t$ for $z_t$ according to $P(l_t|z_t)$, and finally form a latent template $(<z_1,l_1>, \ldots <z_S,l_S>)$. Given a message $X=(x_1,\ldots,x_L)$, the encoder exploits a GRU to transform $X$ into a hidden sequence $H_X=(h_{X,1}, \ldots, h_{X,L})$ with the $i$-th hidden state $h_{X,i} \in \mathbb{R}^{d_2}$ given by
\begin{align*}
    h_{X,i} = \text{GRU}\textsubscript{X}(h_{X,i-1}, e_{x_i}),
\end{align*}
where $e_{x_i} \in \mathbb{R}^{d_3}$ is the embedding of word $x_i$ and $h_{X,0}=0$. Then when predicting the $t$-th word of the response, the decoder calculates the probability $P(y_t|y_{1:t-1}, X, \mathcal{T})$ via
\begin{align*}
        P(y_t|y_{1:t-1}, X, \mathcal{T})= & \text{softmax}(W_2[s_t; c_t]+b_2)\\
        s_t = & \text{GRU}\textsubscript{Y}(s_{t-1}, v_{k}^m).
\end{align*}
with parameters $W_2 \in \mathbb{R}^{V \times 2d_2}$ and $b_2 \in \mathbb{R}^{V}$, $s_t \in \mathbb{R}^{d_2}$ and $s_{t-1} \in \mathbb{R}^{d_2}$ are the hidden states of the decoder for step $t$ and step $t-1$ respectively, $v_k^m$ is defined by Equation (\ref{gatedstate}) where $m$ satisfies $i(m-1) <t \leq i(m)$, $k=t-i(m-1)$, and $o_k^m=\text{GRU}\textsubscript{H}(o_{k-1}^m, [e_{z_m}; e_{y_{t-1}}])$, and $c_t \in \mathbb{R}^{d_2}$ is a context vector of $X$ obtained via attention over $H_X$ \cite{bahdanau2014neural}:
\begin{align*}
    &c_t = \sum^L_{i=1}\alpha_{t,i} h_{X,i},\\
    &\alpha_{t,i} = \frac{\exp(s_{t,i})}{\sum^L_{j=1}\exp(s_{t,j})}\\
    &s_{t,i} = v^\top \tanh (W s_t + U h_i + b),
\end{align*}
where $v, b\in \mathbb{R}^{d_2}$, $W, U \in \mathbb{R}^{d_2 \times d_2}$ are parameters.

\section{Learning Approach}\label{LA}
Intuitively, we can estimate the parameters of the encoder-decoder and fine-tune the parameters of NHSMM by maximizing the likelihood of $\mathcal{D}_P$ (i.e., MLE). However, since $\mathcal{D}_P$ only contains a few pairs, the MLE approach may suffer from a dilemma: (1) if we stop training early, then both the template prior and the encoder-decoder are not sufficiently supervised by the pairs. In that case, the linguistic knowledge in $\mathcal{D}_U$ will play a more important role in response generation and result in irrelevant responses regarding to messages; or (2) if we let the training go deep, then the template prior will be overwhelmed by the pairs in $\mathcal{D}_P$. As a result, the generation model will lose the knowledge obtained from $\mathcal{D}_U$. Since response generation starts from a latent template, we consider learning the model with an adversarial approach \cite{goodfellow2014generative} that can well balance the effect of the latent template and the input message. The learning involves a generator $G$ described in Section \ref{sec:generator} and a discriminator $D$. $G$ is updated with REINFORCE algorithm \cite{williams1992simple} with rewards defined by $D$, and $D$ is updated to distinguish human responses in $\mathcal{D}_P$ from responses generated by $G$. 

\paragraph{Generator Pre-training.} To improve the stability of adversarial learning, we first pre-train $G$ with MLE on $\mathcal{D}_P$. $\forall (X_i,Y_i) \in \mathcal{D}_P$, the template prior $\mathcal{T}_i$ is obtained by running Viterbi algorithm \cite{zucchini2016hidden} on $Y_i$ rather than by sampling. Let $Y_i=(y_{i,1}, \ldots, y_{i,S_i})$, then the objective of pre-training is given by
\begin{equation}
    \label{MLE}
    \frac{1}{n} \sum_{(X_i,Y_i) \in \mathcal{D}_P} \sum_{t=1}^{S_i} \log P(y_{i,t}| y_{i,1:t-1}, X_i, \mathcal{T}_i).
\end{equation}

\paragraph{Discriminator Update.} The discriminator $D$ is defined by a convolutional neural network (CNN) based binary classifier \cite{kim-2014-convolutional}. $D$ takes a message-response pair as input and outputs a score that indicates how likely the response is from humans. In the model, the message and the response are separately embedded as vectors by CNNs, and then the concatenation of the two vectors are fed to a 2-layer MLP to calculate the score. Let $\hat{Y}_i$ be the response generated by $G$ for $X_i$, then $D$ is updated by maximizing the following objective:
\begin{equation}
\label{disc}
  \sum_{(X_i,Y_i) \in \mathcal{D}_P} \log D(X_i,Y_i) + \log (1-D(X_i, \hat{Y}_i))
\end{equation}

\paragraph{Generator Update.} The generator $G$ is updated by the policy gradient method \cite{yu2017seqgan,li2017adversarial}. Let $\hat{y}_{1:t}$ be a partial response generated by $G$ from beam search for message $X$ until step $t$, then we adopt the Monte Carlo (MC) search method and sample $N$ paths that supplement $\hat{y}_{1:t}$ as responses $\{\hat{Y}_i\}_{i=1}^N$. The intermediate reward for $\hat{y}_{1:t}$ is defined as $R_t = \frac{1}{N} \sum^N_{i=1} D(X,\hat{Y}_i)$. The gradient for updating $G$ is given by
\begin{equation}
\label{regs}
    \nabla J(\theta) \approx \sum^S_{t=1} \big[\nabla \log P(\hat{y_t}|\hat{y}_{1:t-1}, X, \mathcal{T}) \cdot R_t \big],
\end{equation}
where $\theta$ represents the parameters of $G$, and $\mathcal{T}$ is a sampled template from the pool $\mathcal{P}$. To control the quality of MC search, we sample from top $50$ most probable words at each step.

The learning algorithm is summarized in Algorithm \ref{alg}. Note that in learning of the generation model from $\mathcal{D}_P$, we freeze the embedding of states (i.e., $e_{z_t}$ in Equation (\ref{gatedstate})) and the embedding of words given by the NHSMM, and update all other parameters in generator pre-training and the following adversarial learning.

\begin{algorithm}[t]
    \caption{Learning a generation model with paired and unpaired data.}\label{alg}
    \begin{algorithmic}[1]
    \small
    \REQUIRE
    NHSMM $H$, generator $G$, discriminator $D$, $\mathcal{D}_U$,  and $\mathcal{D}_P$.
    \STATE
    Initialize $H$, $G$, $D$.
    \STATE
    Learn $H$ from $\mathcal{D}_U$ according to Equation (\ref{backward}).
    \STATE
    Build template pool $\mathcal{P}$ with $\mathcal{D}_U$.
    \STATE
    Pre-train $G$ and $D$.
    \REPEAT
    \FOR{g-steps}
    \STATE
    Sample $(X, Y)$ from $\mathcal{D}_P$.
    \STATE
    Sample a template $\mathcal{T}$ for $(X,Y)$ from $\mathcal{P}$. 
    \STATE
    Generate $\hat{Y}=(\hat{y}_1,\ldots,\hat{y}_S) \sim G(\cdot|X,\mathcal{T})$.
    \FOR{$t$ in $1:S$}
    \STATE
    MC search and compute reward $R_t$ using $D$.
    \ENDFOR
    \STATE
    Update $G$ on $(X, \hat{Y})$ via Equation (\ref{regs}).
    \ENDFOR
    \FOR{d-steps}
    \STATE
    Sample $(X, Y)$ from $\mathcal{D}_P$.
    \STATE
    Sample a template $\mathcal{T}$ for $(X,Y)$ from $\mathcal{P}$. 
    \STATE
    Generate $\hat{Y} \sim G(\cdot|X, \mathcal{T})$ and pair $(X, \hat{Y})$ with $(X, Y)$.
    \STATE
    Update $D$ by Equation ({\ref{disc}}).
    \ENDFOR
    \UNTIL{convergence}
    \end{algorithmic}
    \end{algorithm}
\section{Experiments}
We test the proposed approach on two tasks: question response generation and sentiment response generation. The first task requires a model to generate a question as a response to a given message; while in the second task, as a showcase, responses should express the positive sentiment. 

\subsection{Experiment Setup}
\paragraph{Datasets.} For the question response generation task, we choose the data published in \cite{wang2018learning} as the paired dataset. The data are obtained by filtering $9$ million message-response pairs mined from Weibo with $20$ handcrafted question templates and are split as a training set, a validation set, and a test set with $481$k, $5$k, and $5$k pairs respectively. In addition to the paired data, we crawl $776$k questions from Zhihu, a Chinese community QA website featured by high-quality content, as an unpaired dataset.  Both datasets are tokenized by Stanford Chinese word segmenter\footnote{\url{https://stanfordnlp.github.io/CoreNLP}}. We keep $20,000$ most frequent words in the two data as a vocabulary for the encoder, the decoder, and the NHSMM. The vocabulary covers $95.8$\% words appearing in the messages, in the responses, and in the questions. Other words are replaced with ``UNK''. For the sentiment response generation task, we mine $2$ million message-response pairs from Twitter FireHose, filter responses with the positive sentiment using Stanford Sentiment Annotator toolkit \cite{socher2013recursive}, and obtain $360$k pairs as a paired dataset. As pre-processing, we remove URLs and usernames, and transform each word to its lower case. After that, the data is split as a training set, a validation set, and a test set with $350$k, $5$k, and $5$k pairs respectively. Besides, we extract $1$ million tweets with positive sentiment from a public corpus \cite{cheng2010you} as an unpaired dataset. Top $20,000$ most frequent words in the two data are kept as a vocabulary that covers $99.3$\% words. Words excluded from the vocabulary are treated as ``UNK''.  In both tasks, human responses in the test sets are taken as ground truth for automatic metric calculation. From each test set, we randomly sample $500$ distinct messages and recruit human annotators to judge the quality of responses generated for these messages. 

\paragraph{Evaluation Metrics.} We conduct evaluation with both automatic metrics and human judgements. For automatic evaluation, besides BLEU-1 \cite{papineni2002bleu} and Rouge-L \cite{lin2004rouge}, we follow \cite{serban2017hierarchical} and employ Emebedding Average (Average), Embedding Extrema (Extrema), and Embedding Greedy (Greedy) as metrics. All these metrics are computed by a popular NLG evaluation project available at \url{https://github.com/Maluuba/nlg-eval}. In terms of human evaluation, for each task, we recruit $3$ well-educated native speakers as annotators, and let them compare our model and each of the baselines.  Every time, we show an annotator a message (in total $500$) and two responses, one from our model and the other from a baseline model. Both responses are top $1$ results in beam search, and the two responses are presented in random order.  The annotator then compare the two responses from three aspects: (1) \textbf{Fluency}: if the response is fluent without grammatical error; (2) \textbf{Relevance}: if the response is relevant to the given message; and (3) \textbf{Richness}: if the response contains informative and interesting content, and thus may keep conversation going. For each aspect, if the annotator cannot tell which response is better, he/she is asked to label a ``tie''. Each pair of responses receive $3$ labels on each of the three aspects, and agreements among the annotators are measured by Fleiss' kappa \cite{fleiss1973equivalence}.

\subsection{Baselines}
We compare our model with the following baselines: (1) \textbf{Seq2Seq}: the basic sequence-to-sequence with attention architecture \cite{bahdanau2014neural}. (2) \textbf{CVAE}: the conditional variational autoencoder that represents the relationship between messages and responses with latent variables \cite{zhao2017learning}. We use the code published at \url{https://github.com/snakeztc/NeuralDialog-CVAE}. (3) \textbf{HTD}: the hard typed decoder model proposed in \cite{wang2018learning} that exhibits the best performance on the dataset selected by this work for question response generation. The model estimates distributions over three types of words (i.e., interrogative, topic, and ordinary) and modulates the final distribution during generation. Since our experiments are conducted on the same data as those in \cite{wang2018learning}, we run the code shared at \url{https://github.com/victorywys/Learning2Ask_TypedDecoder} with the default setting. (4) \textbf{ECM}: emotional chatting machine proposed in \cite{zhou2017emotional}. We implement the model with the code published at \url{https://github.com/tuxchow/ecm}. Since the model can handle various emotions, we train the model with the entire $2$ million Twitter message-response pairs labeled with a positive, negative, and neutral sentiment. Thus, when we only focus on responses with positive sentiment, ECM actually performs multi-task learning for response generation. In the test, we set the sentiment label as ``positive''.

We name our model \textbf{S2S-Temp}. Besides the full model, we also examine three variants in order to understand the effect of unpaired data and the role of adversarial learning: (1) S2S-Temp-None. The proposed model is trained only with the paired data, where the NHSMM is estimated from responses in the paired data; (2) S2S-Temp-$50$\%. The proposed model is trained with $50$\% unpaired data; and (3) S2S-Temp-MLE. The pre-trained generator described in Section \ref{LA}. These variants are only involved in automatic evaluation. 

\begin{table}[t!]
    \small
    \centering
    \resizebox{1\linewidth}{!}{
    \begin{tabular}{@{}lc@{\hspace{2mm}}c@{\hspace{2mm}}c@{\hspace{2mm}}c@{\hspace{2mm}}c@{}}
    \toprule
                & BLEU-1  & ROUGE-L & AVERAGE & EXTREMA & GREEDY \\
    \midrule
    Seq2Seq     & 0.037 & 0.111 & 0.656 & 0.438 & 0.456 \\
    CVAE        & 0.094 & 0.088 & 0.685 & 0.414 & 0.422 \\
    HTD         & 0.073 & 0.103 & 0.647 & 0.425 & 0.439 \\
    \midrule
    S2S-Temp-MLE  & 0.097 & 0.119 & 0.699 & 0.438 & 0.457 \\
    S2S-Temp-None & 0.069 & 0.092 & 0.677 & 0.429 & 0.416 \\
    S2S-Temp-$50$\% & 0.091 & 0.113 & 0.702 & 0.442 & 0.461 \\
    S2S-Temp   & \textbf{0.102} & \textbf{0.128} & \textbf{0.710} & \textbf{0.451} & \textbf{0.469} \\
    \bottomrule
    \end{tabular}}
    \caption{Automatic evaluation results for the task of question response generation. Numbers in bold mean that the improvement over the best performing baseline is statistically significant (t-test with $p$-value$<0.01$).}
    \label{tab:qg}
    \end{table}
    
\begin{table}[t!]
    \small
    \centering
    \resizebox{1\linewidth}{!}{
    \begin{tabular}{@{}lc@{\hspace{2mm}}c@{\hspace{2mm}}c@{\hspace{2mm}}c@{\hspace{2mm}}c@{}}
    \toprule
                & BLEU-1  & ROUGE-L & AVERAGE & EXTREMA & GREEDY \\
    \midrule
    Seq2Seq     & 0.065 & 0.118 & 0.726 & 0.474 & 0.582 \\
    CVAE        & 0.088 & 0.081 & 0.727 & 0.408 & 0.563 \\
    ECM         & 0.051 & 0.102 & 0.708 & 0.462 & 0.559 \\
    \midrule
    S2S-Temp-MLE   & 0.103 & 0.124 & 0.732 & 0.458 & 0.593 \\
    S2S-Temp-None  & 0.078 & 0.089 & 0.687 & 0.479 & 0.501 \\
    S2S-Temp-$50$\% & 0.102 & 0.121 & 0.691 & 0.491 & 0.586 \\
    S2S-Temp   & \textbf{0.106} & \textbf{0.130} & \textbf{0.738} & \textbf{0.492} & \textbf{0.603} \\
    \bottomrule
        \end{tabular}}
    \caption{Automatic evaluation results for the task of sentiment response generation. Numbers in bold mean that the improvement over the best performing baseline is statistically significant (t-test, with $p$-value$<0.01$).}
    \label{tab:sentiment}
    \end{table}

\subsection{Implementation Details}
In both tasks, we set the number of states (i.e., $K$) and the the maximal number of emissions (i.e., $D$) in NHSMM as $50$ and $4$ respectively. $d_1$, $d_2$ and $d_3$ are set as $600$, $300$, and $300$ respectively. In adversarial learning, we use three types of filters with window sizes $1$, $2$ and $3$ in the discriminator. The number of filters is $128$ for each type. The number of samples obtained from MC search (i.e., $N$) at each step is $5$. We learn all models using Adam algorithm \cite{kingma2014adam}, monitor perplexity on the validation sets, and terminate training when perplexity gets stable. In our model, learning rates for NHSMM, the generator, and the discriminator are set as \num{1e-3}, \num{1e-5}, and \num{1e-3} respectively. 

\subsection{Evaluation Results}
Table \ref{tab:qg} and Table \ref{tab:sentiment} report the results of automatic evaluation on the two tasks. We can see that on both tasks, S2S-Temp outperforms all baseline models in terms of all metrics, and the improvements are statistically significant (t-test with $p$-value$<0.01$). The results demonstrate that when only limited pairs are available, S2S-Temp can effectively leverage unpaired data to enhance the quality of response generation. Although lacking fine-grained check, from the comparison among S2S-Temp-None, S2S-Temp-$50$\%, and S2S-Temp, we can conclude that the performance of S2S-Temp improves with more unpaired data. Moreover, without unpaired data, our model is even worse than CVAE since the structured templates cannot be accurately estimated from such a few data, and as long as half of the unpaired data are available, the model outperforms the baseline models on most metrics. The results further verified the important role the unpaired data plays in learning of a response generation model from low resources. S2S-Temp is better than S2S-Temp-MLE, indicating that the adversarial learning approach can indeed enhance the relevance of responses regarding to messages.  

\begin{table*}[th!]
    \renewcommand{\arraystretch}{1.3}
        \small
        \centering
        \resizebox{0.9\textwidth}{!}{
        \begin{tabular}{l|ccc|ccc|ccc|c}
            \toprule
            \multirow{2}{*}{\centering Models} & \multicolumn{3}{c|}{Fluency} & \multicolumn{3}{c|}{Relevance} & \multicolumn{3}{c|}{Richness} & Kappa\\
            \cline{2-10}
            & W($\%$) & L($\%$) & T($\%$) & W($\%$) & L($\%$) & T($\%$) & W($\%$) & L($\%$) & T($\%$) &\\
            \midrule
            S2S-Temp vs. Seq2Seq & 20.8 & 18.3 & 60.9 & 30.8 & 22.5 & 46.7 & 42.5 & 19.2 & 38.3 & 0.63\\
            S2S-Temp vs. CVAE & 41.7 & 5.7 & 52.6 & 50.8 & 12.5 & 36.7 & 37.5 & 15.8 & 46.7 & 0.71\\
            S2S-Temp vs. HTD & 35.1 & 19.2 & 45.8 & 30.8 & 25.1 & 44.1 & 37.5 & 30.8 & 31.7 & 0.64\\
             
            \midrule
            S2S-Temp vs. Seq2Seq & 15.6 & 11.5 & 72.9 & 34.4 & 17.2 & 48.4 & 31.9 & 7.3 & 60.8 & 0.68\\
            S2S-Temp vs. CVAE & 48.4 & 9.0 & 42.6 & 31.9 & 5.7 & 62.4 & 31.4 & 8.2 & 60.4 & 0.69\\
            S2S-Temp vs. ECM & 27.1 & 12.3 & 60.6 & 36.9 & 13.9 & 49.2 & 27.9 & 10.6 & 61.5 & 0.78\\
            \bottomrule
        \end{tabular}
        }
        \caption{Human annotation results. W, L, and T refer to Win, Lose and Tie respectively. The first three rows are results on question response generation, and the last three rows are results on sentiment response generation.  The ratios are calculated by combining labels from the three judges.}
        \label{manual_result}
\end{table*}

Table \ref{manual_result} shows the results of human evaluation. In terms of all the three aspects, S2S-Temp is better than all the baseline models. The values of kappa are all above $0.6$, indicating substantial agreement among the annotators. When the size of paired data is small, the basic Seq2Seq model tends to generate more generic responses. That is why the gap between S2S-Temp and Seq2Seq is much smaller on fluency than those on the other two aspects. With the latent variables, CVAE brings both content and noise into responses. Therefore, the gap between S2S-Temp and CVAE is more significant on fluency and relevance than that on richness. HTD can greatly enrich the content of responses, which is consistent with the results in \cite{wang2018learning}, although sometimes the responses might be irrelevant to messages or ill-formed. ECM does not perform well on both automatic evaluation and human judgement.

\subsection{Case Study}
To further understand how S2S-Temp leverages templates for response generation, we show two examples with the test data, one for question response generation in Table \ref{prior_qg} and the other for sentiment response generation in Table \ref{prior_positive}, where subscripts refer to states of the NHSMMs. First, we can see that a template defines a structure for a response. By varying templates, we can have responses with different syntax and semantics for a message. Second, some states may have consistent functions across responses. For example, state $36$ in question response generation may refer to pronouns, and ``I'm'' and ``it was'' correspond to the same state $23$ in sentiment response generation. Finally, some templates provide strong syntactic signals to response generation.  For example, the segmentation of ``Really? I don't believe it'' given by the template (48, 36, 32) matches the parsing result ``FRAG + LS + VP '' given by stanford syntactic parser.

\begin{CJK*}{UTF8}{gkai} 
\begin{table}[t!]
    \small
    \centering
    \resizebox{1\linewidth}{!}{
    \begin{tabular}{l m{6.0cm}}
        \toprule
        {\textbf{Message:}} & 真的假的？我瘦了16斤 \\
                            & Really? I lost 17.6 pounds \\
        \midrule
        \textbf{Responses:} & [你]$_{36}$ \enspace [瘦\enspace了]$_{31}$ \enspace [吗\enspace?]$_{13}$\\
                           & [You]$_{36}$ \enspace [lost weight]$_{31}$\enspace [?]$_{13}$\\[0.2cm]
                           & [你\enspace是\enspace怎么]$_{14}$ \enspace [做到\enspace的]$_{42}$ \enspace [?]$_{26}$\\
                           & [How do you]$_{14}$ \enspace [make it]$_{42}$ \enspace [?]$_{26}$\\[0.2cm]
                           & [真的\enspace吗\enspace?]$_{48}$ \enspace [我]$_{36}$ \enspace [不\enspace信]$_{32}$\\
                           & [Really ?]$_{48}$ \enspace [I]$_{36}$ \enspace [don't believe it]$_{32}$\\
        \bottomrule
    \end{tabular}}
    \caption{Question response generation with various templates.}
    \label{prior_qg}
\end{table}

\begin{table}[t!]
    \normalsize
    \centering
    \resizebox{1\linewidth}{!}{
    \begin{tabular}{l m{7cm}}
        \toprule
        {\textbf{Message:}} & One of my favoriate  Eddie Murphy movies! \\
        \midrule
        \textbf{Responses:} & [it\enspace 's]$_{29}$ \enspace [a\enspace brilliant]$_{35}$ \enspace [movie]$_{37}$\\[0.15cm]
                           & [i \enspace screamed]$_{29}$ \enspace [! ! !]$_{1}$ \\[0.15cm]
                           & [it\enspace was]$_{23}$ \enspace [so\enspace underrated]$_{14}$ \enspace [!]$_{31}$\\[0.15cm]
                          & [honestly\enspace .]$_{30}$ [i\enspace 'm]$_{23}$ [so\enspace pumped]$_{14}$ [to\enspace watch \enspace it]$_{31}$\\[0.15cm]
                           & [yeah\enspace ,]$_{47}$ \enspace [i \enspace was \enspace watching]$_{48}$\\[0.15cm]
    
        \bottomrule
    \end{tabular}
    }
    \caption{Sentiment response generation with various templates.}
    \label{prior_positive}
\end{table}
\end{CJK*}

\section{Conclusions}
We study low-resource response generation for open domain dialogue systems by assuming that paired data are insufficient for modeling the relationship between messages and responses. To augment the paired data, we consider transferring knowledge from unpaired data to response generation through latent templates parameterized as a hidden semi-markov model, and take the templates as prior in generation. Evaluation results on question response generation and sentiment response generation indicate that when limited pairs are available, our model can significantly outperform several state-of-the-art response generation models. 

\section*{Acknowledgement}
We appreciate the valuable comments provided by the anonymous reviewers. This work is supported in part by the National Natural Science Foundation of China (Grand Nos. U1636211, 61672081, 61370126), and the National Key R\&D Program of China (No. 2016QY04W0802).

\bibliography{emnlp-ijcnlp-2019}
\bibliographystyle{acl_natbib}

\end{document}